\documentclass[review]{elsarticle}
\usepackage{amsmath}
\usepackage{indentfirst}
\usepackage{booktabs}
\usepackage{subfigure}
\usepackage{epsfig}
\usepackage{amssymb}
\usepackage{graphicx}
\usepackage{float}
\usepackage{setspace}
\usepackage{tabularx,booktabs}
\usepackage{url}
\biboptions{numbers,sort&compress}
\journal{Artificial Intelligence In Medicine}
\usepackage{multirow}
\usepackage{rotating}
\usepackage{sverb, longtable}

\begin{document}
\begin{spacing}{1.5}
\begin{frontmatter}

\title{Quantum dynamical mode (QDM): A possible extension of belief function
}

\author[address1]{Fuyuan Xiao\corref{label1}}
\address[address1]{School of Computer and Information Science, Southwest University, China, \\ No.2 Tiansheng Road, BeiBei District, Chongqing, 400715,P.R.China}
\cortext[label1]{Corresponding author at: School of Computer and Information Science, Southwest University, No.2 Tiansheng Road, BeiBei District, Chongqing, 400715, China. E-mail: xiaofuyuan@swu.edu.cn}

\begin{abstract}
Dempster--Shafer evidence theory has been widely used in various fields of applications, because of the flexibility and effectiveness in modeling uncertainties without prior information.
Besides, it has been proven that the quantum theory has powerful capabilities of solving the decision making problems, especially for modelling human decision and cognition.
However, due to the inconsistency of the expression, the classical Dempster--Shafer evidence theory modelled by real numbers cannot be integrated directly with the quantum theory modelled by complex numbers.
So, how can we establish a bridge of communications between the classical Dempster--Shafer evidence theory and the quantum theory?
To answer this question, a generalized Dempster--Shafer evidence theory is proposed in this paper.
The main contribution in this study is that, unlike the existing evidence theory, a mass function in the generalized Dempster--Shafer evidence theory is modelled by a complex number, called as a complex mass function.
In addition, compared with the classical Dempster's combination rule, the condition in terms of the conflict coefficient between two evidences $\mathbf{K} < 1$ is released in the generalized Dempster's combination rule so that it is more general and applicable than the classical Dempster's combination rule.
When the complex mass function is degenerated from complex numbers to real numbers, the generalized Dempster's combination rule degenerates to the classical evidence theory under the condition that the conflict coefficient between the evidences $\mathbf{K}$ is less than 1.
This generalized Dempster--Shafer evidence theory provides a promising way to model and handle more uncertain information.
Numerical examples are illustrated to show the efficiency of the generalized Dempster--Shafer evidence theory.
Finally, an application of an evidential quantum dynamical model is implemented by integrating the generalized Dempster--Shafer evidence theory with the quantum dynamical model.
From the experimental results, it validates the feasibility and effectiveness of the proposed method.
\end{abstract}


\end{frontmatter}


\section{Introduction}\label{Introduction}
How to measure the uncertainty has been an attracting issue in information fusion area.
The amount of theories had been proposed and extended for measuring the uncertainty, including
the rough sets theory~\cite{walczak1999rough},
fuzzy~sets~theory~\cite{zadeh1965fuzzy,liu2013fuzzy,Den2011Maximum,zhengrong2017ADAC},
evidence~theory~\cite{Dempster1967Upper,Jiang2017mGCR,xuhonghui2018,XDWJIJIS21929},
Z~numbers~\cite{zadeh2011note,Kang2017Stable},
D~numbers~\cite{zhangqi2017,xiao2016intelligent,DAHPcredibility2018,liubaoyuIJCCC2017,Bian2018Failure,zhou2017dependence},
evidential~reasoning~\cite{yang2012belief,fu2015group,yang2014interactive,yang2013evidential},
and so on~\cite{ma2012qualitative,Cuzzolin2014Learning}.

As an uncertainty reasoning tool, Dempster--Shafer evidence theory was firstly presented by Dempster~\cite{Dempster1967Upper} in 1967 year.
Soon afterwards, it had been developed by Shafer~\cite{shafer1976mathematical} in 1976 year.
Thanks to the flexibility and effectiveness in modeling uncertainties without prior information, Dempster--Shafer evidence theory has been widely used in various fields of applications,
like decision making~\cite{jiang2018IJSS,feiliguo2017new,jiang2017Intuitionistic,jiang2017evidence,deng2016evidenceIEEE,Wang2017IJCCC},
pattern recognition~\cite{denoeux1995k,ma2016evidential,liu2016adaptive,Liu2018Change},
risk analysis~\cite{dutta2015uncertainty,zhengxianglin2017,zhang2017improved},
supplier selection~\cite{liuDEMATEL2017},
fault diagnosis~\cite{Jiang2017FMEA,fan2006fault,jiang2017failure,Du2016FMEA},
and so on~\cite{zheng2017evaluation,Liu2017Classifier,kangbingyi2017,Jiang2016CAIE,dong2017location,Den2014Optimal,zheng2017fuzzy}.
Although Dempster--Shafer evidence theory is a very useful uncertainty reasoning tool, the fusing of highly conflicting evidences may result in counter-intuitive results~\cite{zadeh1986simple}.
To address this issue, two main kinds of methodologies have been studied~\cite{lefevre2002belief,Jiang2017Ordered,ma2015belief}.
One methodology focus on modifying Dempster's combination rule~\cite{smets1990combination,dubois1988representation,yager1987dempster}, while the other one focus on pre-processing the bodies of evidences~\cite{murphy2000combining,zhang2014novel}.

Currently, the quantum theory has became an interesting and hot topic in solving the decision making problems.
As justified in literatures~\cite{Pothos2009A,busemeyer2012quantum,Bruza2015Quantum}, the quantum theory can better describe the way humans make judgments towards uncertainty and decisions under conflict environment.
It has been known that the quantum theory is represented by complex probability~\cite{ablowitz2003complex}.
So the question remains, can we leverage the complex probability to express the
Dempster--Shafer evidence theory in the same way?
As a pioneer, Deng~\cite{deng2017mess} first proposed a meta mass function expressed by complex numbers in Dempster--Shafer evidence theory.
Inspired by his research work, a generalized Dempster--Shafer evidence theory is proposed in this study.
The proposed method is both orthogonal and complementary to Deng~\cite{deng2017mess}'s method.
Specifically, a mass function in the generalized Dempster--Shafer evidence theory is modelled by a complex number, called as a complex mass function.
Furthermore, compared with the classical Dempster's combination rule, the condition in terms of the conflict coefficient between two evidences $\mathbf{K} < 1$ is released in the generalized Dempster's combination rule.
Hence, the proposed method is more general and applicable than the classical Dempster's combination rule.
In particular, when the complex mass function is degenerated from complex numbers to real numbers, the generalized Dempster's combination rule degenerates to the classical evidence theory under the condition that the conflict coefficient between two evidences $\mathbf{K}$ is less than 1.
In this context, the generalized Dempster--Shafer evidence theory provides a promising way to model and handle more uncertain information.
Consequently, several numerical examples are provided to illustrate the efficiency of the generalized Dempster--Shafer evidence theory.
Besides, an application of an evidential quantum dynamical model is implemented by integrating the generalized Dempster--Shafer evidence theory with the quantum dynamical model.
The experimental results validate the feasibility and effectiveness of the proposed method.

The remaining content of this paper is organised below.
Section~\ref{Preliminaries} introduces the preliminaries of this paper briefly.
In Section~\ref{Proposed method}, a generalized Dempster--Shafer evidence theory is proposed.
Section~\ref{Experiments} gives numerical examples to illustrate the effectiveness of the proposal.
In Section~\ref{Application}, an application of an evidential quantum dynamical model is implemented.
Finally, Section~\ref{Conclusion} gives the conclusion.

\section{Preliminaries}\label{Preliminaries}

\subsection{Complex number~\cite{ablowitz2003complex,deng2017mess}}\label{Complexnumber}
A complex number $z$ is a number of the form,
\begin{equation}\label{eq_complexnumber}
z=x + yi,
\end{equation}
where $x$ and $y$ are real numbers and $i$ is the imaginary unit, satisfying $i^2 = -1$.

Give two complex numbers $z_1=x_1 + y_1i$ and $z_2=x_2 + y_2i$, the addition is defined as follows:
\begin{equation}\label{eq_addition}
z_1+z_2=(x_1 + y_1i)+(x_2 + y_2i)=(x_1+x_2)+(y_1+y_2)i.
\end{equation}

The subtraction is defined as follows:
\begin{equation}\label{eq_subtraction}
z_1-z_2=(x_1 + y_1i)-(x_2 + y_2i)=(x_1-x_2)+(y_1-y_2)i.
\end{equation}

The multiplication is defined as follows:
\begin{equation}\label{eq_multiplication}
(x_1 + y_1i)(x_2 + y_2i)=(x_1x_2-y_1y_2)+(x_1y_2+x_2y_1)i.
\end{equation}

The division is defined as follows:
\begin{equation}\label{eq_division}
\frac{x_1 + y_1i}{x_2 + y_2i}=\frac{x_1x_2+y_1y_2}{x_2^2+y_2^2}+\frac{x_2y_1-x_1y_2}{x_2^2+y_2^2}i.
\end{equation}

An important parameter is the absolute value (or modulus or magnitude) of a complex number $z = x + yi$ is
\begin{equation}\label{eq_magnitude}
r=|z|=\sqrt{x^2+y^2},
\end{equation}
where if $z$ is a real number (i.e., $y = 0$), then $r = |x|$.

The square of the absolute value is
\begin{equation}\label{eq_squaremagnitude}
|z|^2=z\bar{z}=x^2+y^2,
\end{equation}
where $\bar{z}$ is the complex conjugate of $z$, i.e., $\bar{z}=x - yi$.

\subsection{Dempster--Shafer evidence theory~\cite{Dempster1967Upper,shafer1976mathematical}}
Dempster--Shafer evidence theory is extensively applied to handle uncertain information that belongs to the category of artificial intelligence.
Because Dempster--Shafer evidence theory is flexible and effective in modeling the uncertainty regardless of prior information, it requires weaker conditions compared with the Bayesian theory of probability.
When the probability is~confirmed, Dempster--Shafer evidence theory degenerates to the probability theory and is considered as a generalization of Bayesian inference.
In addition, Dempster--Shafer evidence theory has the advantage that it can directly express the ``uncertainty'' via allocating the probability into the set's subsets, which consists of multi-objects, instead of a single object.
Furthermore, it is capable of combining the bodies of evidence to derive new evidence.
The basic concepts and definitions are described as~below.

\newtheorem{myDef}{Definition}
\begin{myDef}(Frame of discernment)

Let $\Theta$ be a nonempty set of events that are mutually-exclusive and collectively-exhaustive, defined by:
\begin{equation}\label{eq_Frameofdiscernment1}
 \Theta = \{F_{1}, F_{2}, \ldots, F_{i}, \ldots, F_{N}\},
\end{equation}
in which the set $\Theta$ denotes a frame of discernment.

The power set of $\Theta$ is represented as $2^{\Theta}$, where:
\begin{equation}\label{eq_Frameofdiscernment2}
 2^{\Theta} = \{\emptyset, \{F_{1}\}, \{F_{2}\}, \ldots, \{F_{N}\}, \{F_{1}, F_{2}\}, \ldots, \{F_{1}, F_{2}, \ldots, F_{i}\}, \ldots, \Theta\},
\end{equation}
and $\emptyset$ is an empty set.

When $A$ is an element of the power set of $\Theta$, i.e., $A \in 2^{\Theta}$, $A$ is called a hypothesis or proposition.
\end{myDef}

\begin{myDef}(Mass function)

In the frame of discernment $\Theta$, a mass function $m$ is represented as a mapping from $2^{\Theta}$ to [0, 1] that is defined as:
\begin{equation}\label{eq_Massfunction1}
 m: \quad 2^{\Theta} \rightarrow [0, 1],
\end{equation}
which meets the conditions below:
\begin{equation}
\label{eq_Massfunction2}
\begin{aligned}
 m(\emptyset) &= 0, \\
 \sum\limits_{A \in 2^{\Theta}} m(A) &= 1.
 \end{aligned}
\end{equation}
\end{myDef}

The mass function $m$ in the Dempster--Shafer evidence theory can also be called a basic belief assignment (BBA).
When $m(A)$ is greater than zero, $A$ as the element of $2^{\Theta}$ is named as a focal element of the mass function, where the mass function $m(A)$ indicates how strongly the evidence supports the proposition or hypothesis $A$.

\begin{myDef}(Belief function)

Let $A$ be a proposition where $A \subseteq \Theta$; the belief function $Bel$ of the proposition $A$ is defined by:
\begin{equation}\label{eq_belieffunction}
\begin{aligned}
Bel: \quad &2^{\Theta} \rightarrow [0, 1], \\
&Bel(A) = \sum\limits_{B \subseteq A} m(B).
\end{aligned}
\end{equation}

The plausibility function $Pl$ of the proposition $A$ is defined by:
\begin{equation}\label{eq_plausibilityfunction}
\begin{aligned}
Pl: \quad &2^{\Theta} \rightarrow [0, 1], \\
&Pl(A) = 1 - Bel(\bar{A}) = \sum\limits_{B \cap A \neq \emptyset} m(B),
\end{aligned}
\end{equation}
where $\bar{A}$ is the complement of $A$, such that $\bar{A} = \Theta - A$.
\end{myDef}

Apparently, the plausibility function $Pl(A)$ is equal to or greater than the belief function $Bel(A)$, where the belief function $Bel$ is the lower limit function of the proposition $A$, and the plausibility function $Pl$ is the upper limit function of the proposition $A$.

\begin{myDef}(Dempster's rule of combination)

Let two basic belief assignments (BBAs) be $m_1$ and $m_2$ on the frame of discernment $\Theta$ where the BBAs $m_1$ and $m_2$ are independent; Dempster's rule of combination, defined by $m = m_1 \oplus m_2$, which is called the orthogonal sum, is represented as below:
\begin{equation}\label{eq_Dempsterrule1}
{
m(A) = \left\{ \begin{array}{l}
\begin{array}{*{20}{c}}
{\frac{1}{1-K} \sum\limits_{B \cap C = A} m_1(B) m_2(C),}&{{\kern 30pt} A \neq \emptyset,}
\end{array}\\
\begin{array}{*{20}{c}}
{0,}&{{\kern 136pt} A = \emptyset,}
\end{array}
\end{array} \right.
}\end{equation}
with
\begin{equation}\label{eq_Dempsterrule2}
{
K = \sum\limits_{B \cap C = \emptyset} m_1(B) m_2(C),
}\end{equation}
where $B$ and $C$ are also the elements of $2^{\Theta}$ and $K$ is a constant that presents the conflict coefficient between the BBAs $m_1$ and $m_2$.
\end{myDef}

Notice that Dempster's combination rule is only practicable for the BBAs $m_1$ and $m_2$ under the condition that $K < 1$.

\begin{myDef}(Pignistic probability transformation)

Let $m$ be a basic belief assignment on the frame of discernment $\Theta$ and $A$ be a proposition where $A \subseteq \Theta$, the pignistic probability transformation function is defined by
\begin{equation}\label{eq_Pignistic}
Bet(B) = \sum_{B \subseteq A} \frac{m(A)}{|A|},
\end{equation}
\end{myDef}
where $|A|$ represents the cardinality of $A$.

\section{Generalized Dempster--Shafer evidence theory}\label{Proposed method}

Let $\Omega$ be a nonempty set of events that are mutually-exclusive and collectively-exhaustive, defined by:
\begin{equation}\label{eq_Frameofdiscernment1}
 \Omega = \{E_{1}, E_{2}, \ldots, E_{i}, \ldots, E_{N}\},
\end{equation}
in which the set $\Omega$ denotes a frame of discernment.

The power set of $\Omega$ is represented as $2^{\Omega}$, where:
\begin{equation}\label{eq_Frameofdiscernment2}
 2^{\Omega} = \{\emptyset, \{E_{1}\}, \{E_{2}\}, \ldots, \{E_{N}\}, \{E_{1}, E_{2}\}, \ldots, \{E_{1}, E_{2}, \ldots, E_{i}\}, \ldots, \Omega\},
\end{equation}
and $\emptyset$ is an empty set.

\begin{myDef}(Complex mass function)\label{def_Complexmassfunction}

In the frame of discernment $\Omega$, a complex mass function $\mathbf{m}$ is modelled as a complex number: 
\begin{equation}\label{eq_GMassfunction}
 x+yi,
\end{equation}
with
\begin{equation}\label{eq_GMassfunction}
x^2+y^2 \in [0, 1],
\end{equation}
and is represented as a mapping from $2^{\Omega}$ to $\mathbb{C}$, denoted as:
\begin{equation}\label{eq_GMassfunction1}
 \mathbf{m}: \quad 2^{\Omega} \rightarrow \mathbb{C},
\end{equation}
which meets the conditions below:
\begin{equation}
\label{eq_Massfunction2}
\begin{aligned}
 \mathbf{m}(\emptyset) &= 0, \\
 \sum\limits_{A \in 2^{\Omega}} \mathbf{m}(A) &= 1.
 \end{aligned}
\end{equation}
\end{myDef}

The complex mass function $\mathbf{m}$ modelled as a complex number in the generalized Dempster--Shafer evidence theory can also be called a complex basic belief assignment (CBBA).
When $|\mathbf{m}(A)|$ is greater than zero, $A$ as the element of $2^{\Omega}$ is named as a focal element of the generalized mass function, where the mass function $|\mathbf{m}(A)|$ indicates how strongly the evidence supports the proposition or hypothesis $A$.
Note that $|\mathbf{m}(A)|$ is the magnitude of $\mathbf{m}(A)$ which can be calculated based on Eq.~(\ref{eq_magnitude}).

\begin{myDef}(Complex belief function)

Let $A$ be a proposition where $A \subseteq \Omega$; a complex belief function $Bel_c$ of the proposition $A$ is defined by:
\begin{equation}\label{eq_Gbelieffunction}
\begin{aligned}
Bel_c: \quad &2^{\Omega} \rightarrow \mathbb{C}, \\
&Bel_c(A) = |\sum\limits_{B \subseteq A} \mathbf{m}(B)|.
\end{aligned}
\end{equation}

The complex plausibility function $Pl_c$ of the proposition $A$ is defined by:
\begin{equation}\label{eq_Gplausibilityfunction}
\begin{aligned}
Pl_c: \quad &2^{\Omega} \rightarrow \mathbb{C}, \\
&Pl_c(A) = 1 - |Bel_c(\bar{A})| = |\sum\limits_{B \cap A \neq \emptyset} \mathbf{m}(B)|,
\end{aligned}
\end{equation}
where $\bar{A}$ is the complement of $A$, such that $\bar{A} = \Omega - A$.
\end{myDef}

Apparently, the plausibility function $Pl_c(A)$ is equal to or greater than the belief function $Bel_c(A)$, where the belief function $Bel_c$ is the lower limit function of the proposition $A$, and the plausibility function $Pl_c$ is the upper limit function of the proposition $A$.

\begin{myDef}(Generalized Dempster's rule of combination)

Let two complex basic belief assignments (CBBAs) be $\mathbf{m}_1$ and $\mathbf{m}_2$ on the frame of discernment $\Omega$ where the CBBAs $\mathbf{m}_1$ and $\mathbf{m}_2$ are independent; the generalized Dempster's rule of combination, defined by $\mathbf{m} = \mathbf{m}_1 \oplus \mathbf{m}_2$, which is called the orthogonal sum, is represented as below:
\begin{equation}\label{eq_GDempsterrule1}
{
\mathbf{m}(A) = \left\{ \begin{array}{l}
\begin{array}{*{20}{c}}
{\frac{1}{1-\mathbf{K}} \sum\limits_{B \cap C = A} \mathbf{m}_1(B) \mathbf{m}_2(C),}&{{\kern 30pt} A \neq \emptyset,}
\end{array}\\
\begin{array}{*{20}{c}}
{0,}&{{\kern 138pt} A = \emptyset,}
\end{array}
\end{array} \right.
}\end{equation}
with
\begin{equation}\label{eq_GDempsterrule2}
{
\mathbf{K} = \sum\limits_{B \cap C = \emptyset} \mathbf{m}_1(B) \mathbf{m}_2(C),
}\end{equation}
where $B$ and $C$ are also the elements of $2^{\Omega}$ and $\mathbf{K}$ is a constant that presents the conflict coefficient between the CBBAs $\mathbf{m}_1$ and $\mathbf{m}_2$.
\end{myDef}

\newtheorem{Remark}{Remark}
\begin{Remark}
Generalized Dempster's combination rule is only practicable for the CBBAs $\mathbf{m}_1$ and $\mathbf{m}_2$ under the condition that the conflict coefficient $\mathbf{K} \neq 1$.
\end{Remark}

\begin{Remark}
Compared with the classical Dempster's combination rule, the condition in terms of the conflict coefficient $\mathbf{K} < 1$ is released in the generalized Dempster's combination rule so that it is more general and applicable than the classical Dempster's combination rule.
\end{Remark}

\begin{Remark}
When the complex mass function is degenerated from complex numbers to real numbers, the generalized Dempster's combination rule degenerates to the classical evidence theory under the condition that the conflict coefficient $\mathbf{K} < 1$.
\end{Remark}

An example is given to prove that the condition $\mathbf{K} < 1$ is ignored in the generalized Dempster's combination rule and depicts the variation of the magnitude of conflict coefficient $|\mathbf{K}|$ between two CBBAs where $|\mathbf{K}|$ can be calculated based on Eq.~(\ref{eq_magnitude}).

\newtheorem{exmp}{Example}
\begin{exmp}\label{exa_conflictcoefficient}
\rm Supposing that there are two CBBAs $\mathbf{m}_1$ and $\mathbf{m}_2$ in the frame of discernment $\Omega=\{A,B\}$, and the two CBBAs are given as follows:
\end{exmp}

\begin{tabular}[t]{l}
$m_1:$ $m_1(A)=x+yi$, $m_1(B)=1-x-yi$;\\
$m_2:$ $m_2(A)=0.5+0.5i$, $m_2(B)=0.5-0.5i$.\\
\specialrule{0em}{6pt}{6pt}
\end{tabular}

According to Definition~\ref{def_Complexmassfunction}, the parameters $x$ and $y$ are set within [-1, 1] satisfying the conditions that $x^2+y^2 \in [0, 1]$ and $(1-x)^2+y^2 \in [0, 1]$ at the same time.

\begin{figure*}[htbp]
\centering
\subfigure[]{
\label{conflictcoefficient:a} 
\includegraphics[width=.7\linewidth,clip]{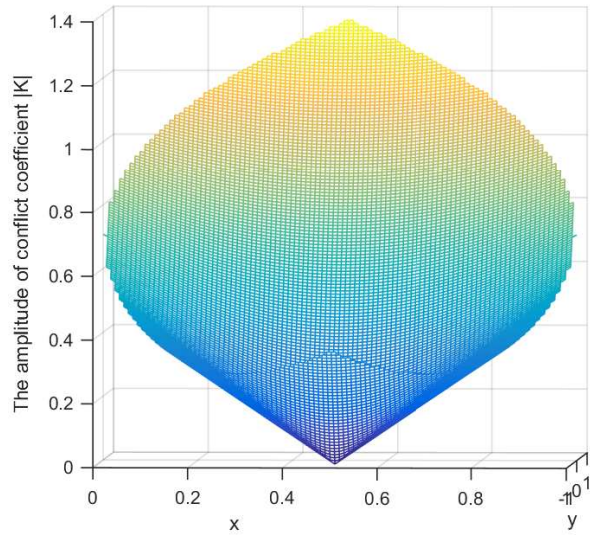}
}
\subfigure[]{
\label{conflictcoefficient:b} 
\centering
\includegraphics[width=.7\linewidth,clip]{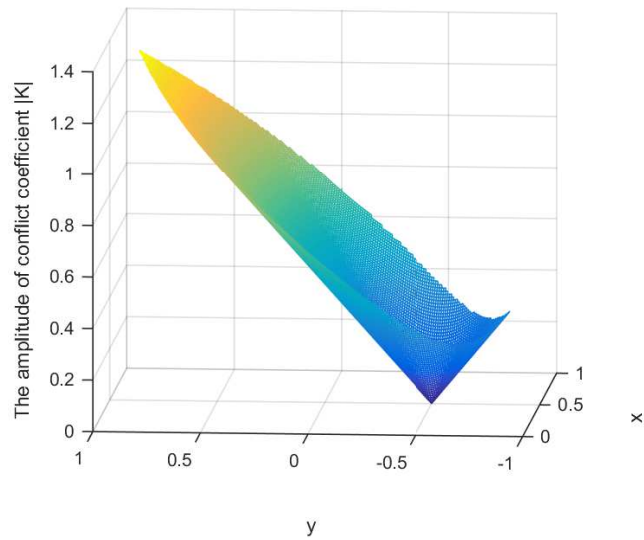}
}
\caption{An example of the variation of $|\mathbf{K}|$ between two CBBAs from the front and the side angles.}
\label{conflictcoefficient12} 
\end{figure*}

Fig.~\ref{conflictcoefficient12} show the results of the magnitude of conflict coefficient $|\mathbf{K}|$ between the two CBBAs $m_1$ and $m_2$ from different angles.

In particular, as shown in Fig.~\ref{conflictcoefficient12}, in the case that $x=1$ and $y=0$, we can obtain the $m_1(A)=1$ and $m_1(B)=0$.
The conflict coefficient $\mathbf{K}$ is calculated as $1\times(0.5-0.5i)+0\times(0.5+0.5i)$; then the magnitude of conflict coefficient $|\mathbf{K}|$ between the two CBBAs $m_1$ and $m_2$ is 0.7071.

When $x=0$ and $y=0$, the $m_1(A)=0$ and $m_1(B)=1$ can be obtained.
The conflict coefficient $\mathbf{K}$ is calculated as $0\times(0.5-0.5i)+1\times(0.5+0.5i)$; then the magnitude of conflict coefficient  $|\mathbf{K}|$ between the two CBBAs $m_1$ and $m_2$ is 0.7071 which shows the same result as the case that $x=1$ and $y=0$.

In the case that $x=0.5$ and $y=-0.5$, we can obtain the $m_1(A)=0.5-0.5i$ and $m_1(B)=0.5+0.5i$.
The conflict coefficient $\mathbf{K}$ is calculated as $(0.5-0.5i)\times(0.5-0.5i)+(0.5+0.5i)\times(0.5+0.5i)$; then the magnitude of conflict coefficient $|\mathbf{K}|$ between the two CBBAs $m_1$ and $m_2$ is 0.

When $x=0.5$ and $y=0.8660$, the $m_1(A)=0.5+0.8660i$ and $m_1(B)=0.5-0.8660i$ can be calculated.
The conflict coefficient $\mathbf{K}$ is calculated as $(0.5+0.8660i)\times(0.5-0.5i)+(0.5-0.8660i)\times(0.5+0.5i)$; then the magnitude of conflict coefficient  $|\mathbf{K}|$ between the two CBBAs $m_1$ and $m_2$ is 1.3660.

In the case that $x=0.5$ and $y=-0.8660$, the $m_1(A)=0.5-0.8660i$ and $m_1(B)=0.5+0.8660i$ can be calculated.
The conflict coefficient $\mathbf{K}$ is calculated as $(0.5-0.8660i)\times(0.5-0.5i)+(0.5+0.8660i)\times(0.5+0.5i)$; then the magnitude of conflict coefficient  $|\mathbf{K}|$ between the two CBBAs $m_1$ and $m_2$ is 0.3660.

\begin{myDef}(Complex pignistic probability transformation)

Let $\mathbf{m}$ be a complex basic belief assignment on the frame of discernment $\Omega$ and $A$ be a proposition where $A \subseteq \Omega$, the complex pignistic probability transformation function is defined by
\begin{equation}\label{eq_pignistic}
Bet_c(B) = \sum_{B \subseteq A} \frac{\mathbf{m}(A)}{|A|},
\end{equation}
\end{myDef}
where $|A|$ represents the cardinality of $A$.

\section{Numerical examples}\label{Experiments}
In this section, several numerical examples are illustrated to show the effectiveness of the generalized Dempster--Shafer evidence theory.

\begin{exmp}\label{exa_1}
\rm Supposing that there are two CBBAs $\mathbf{m}_1$ and $\mathbf{m}_2$ in the frame of discernment $\Omega=\{A,B\}$, and the two CBBAs are given as follows:
\end{exmp}

\begin{tabular}[t]{l}
$\mathbf{m}_1:$
$\mathbf{m}_1(A)=0.1-\frac{\sqrt{2}}{8}i$, $\mathbf{m}_1(B)=0.7+\frac{2\sqrt{2}}{8}i$, $\mathbf{m}_1(A,B)=0.2-\frac{\sqrt{2}}{8}i$;\\
$\mathbf{m}_2:$
$\mathbf{m}_2(A)=0.1+\frac{2\sqrt{3}}{10}i$, $\mathbf{m}_2(B)=0.6+\frac{\sqrt{3}}{10}i$, $\mathbf{m}_2(A,B)=0.3-\frac{3\sqrt{3}}{10}i$.\\
\specialrule{0em}{6pt}{6pt}
\end{tabular}

Then, the fusing results are calculated by utilising Eq.~(\ref{eq_GDempsterrule1}) as follows:

\begin{tabular}[t]{l}
$\mathbf{m}(A)$ = 0.0979 + 0.0186i, \\
$\mathbf{m}(B)$ = 0.9031 - 0.1820i, \\
$\mathbf{m}(A,B)$ = -0.0010 + 0.1634i. \\
\specialrule{0em}{6pt}{6pt}
\end{tabular}

It is verified that $\mathbf{m}(A)$ + $\mathbf{m}(B)$ + $\mathbf{m}(A,B)$ = 1 in this example.

\begin{exmp}\label{exa_2}
\rm Supposing that there are two CBBAs $\mathbf{m}_1$ and $\mathbf{m}_2$ in the frame of discernment $\Omega=\{A,B\}$, and the two CBBAs are given as follows:
\end{exmp}

\begin{tabular}[t]{l}
$\mathbf{m}_1:$
$\mathbf{m}_1(A)=0.1+\frac{2\sqrt{3}}{10}i$, $\mathbf{m}_1(B)=0.6+\frac{\sqrt{3}}{10}i$, $\mathbf{m}_1(A,B)=0.3-\frac{3\sqrt{3}}{10}i$.\\
$\mathbf{m}_2:$
$\mathbf{m}_2(A)=0.1-\frac{\sqrt{2}}{8}i$, $\mathbf{m}_2(B)=0.7+\frac{2\sqrt{2}}{8}i$, $\mathbf{m}_2(A,B)=0.2-\frac{\sqrt{2}}{8}i$;\\
\specialrule{0em}{6pt}{6pt}
\end{tabular}

The fusing results by utilising Eq.~(\ref{eq_GDempsterrule1}) are calculated as follows:

\begin{tabular}[t]{l}
$\mathbf{m}(A)$ = 0.0979 + 0.0186i, \\
$\mathbf{m}(B)$ = 0.9031 - 0.1820i, \\
$\mathbf{m}(A,B)$ = -0.0010 + 0.1634i. \\
\specialrule{0em}{6pt}{6pt}
\end{tabular}

It is obvious that $\mathbf{m}(A)$ + $\mathbf{m}(B)$ + $\mathbf{m}(A,B)$ = 1 in this example.

Through Example~\ref{exa_1} and Example~\ref{exa_2}, it proves that the generalized Dempster--Shafer evidence theory satisfies the commutative law.

\begin{exmp}\label{exa_3}
\rm Supposing that there are two CBBAs $\mathbf{m}_1$ and $\mathbf{m}_2$ in the frame of discernment $\Omega=\{A,B\}$ where they are degenerated to real numbers, and the two CBBAs are given as follows:
\end{exmp}

\begin{tabular}[t]{l}
$\mathbf{m}_1:$
$\mathbf{m}_1(A)=0.8$,
$\mathbf{m}_1(B)=0.2$;\\
$\mathbf{m}_2:$
$\mathbf{m}_2(A)=0.9$,
$\mathbf{m}_2(B)=0.1$.\\
\specialrule{0em}{6pt}{6pt}
\end{tabular}

On the one hand, by utilising Eq.~(\ref{eq_GDempsterrule1}) of the generalized Dempster's rule of combination, the fusing results are generated as follows:

\begin{tabular}[t]{l}
$\mathbf{m}_1(A)=0.9730$,\\
$\mathbf{m}_1(B)=0.0270$;\\
\specialrule{0em}{6pt}{6pt}
\end{tabular}

On the other hand, based on Eq.~(\ref{eq_Dempsterrule1}) of the classical Dempster's rule of combination, the fusing results are calculated as follows:

\begin{tabular}[t]{l}
$\mathbf{m}_1(A)=0.9730$,\\
$\mathbf{m}_1(B)=0.0270$;\\
\specialrule{0em}{6pt}{6pt}
\end{tabular}

It is easy to see that the fusing results from the generalized Dempster's rule of combination is exactly the same as the fusing results from the classical Dempster's rule of combination.
In this example, the conflict coefficient $\mathbf{K}$ is 0.2600.

This example proves that when the complex mass function is degenerated from complex numbers to real numbers, the generalized Dempster's combination rule degenerates to the classical evidence theory under the condition that the conflict coefficient between the evidences $\mathbf{K}$ is less than 1.

\begin{exmp}\label{exa_4}
\rm Supposing that there are two highly conflicting CBBAs $\mathbf{m}_1$ and $\mathbf{m}_2$ in the frame of discernment $\Omega=\{A,B,C\}$ where they are degenerated to real numbers, and the two CBBAs are given as follows:
\end{exmp}

\begin{tabular}[t]{l}
$\mathbf{m}_1:$
$\mathbf{m}_1(A)=0.99$,
$\mathbf{m}_1(C)=0.01$;\\
$\mathbf{m}_2:$
$\mathbf{m}_2(B)=0.99$,
$\mathbf{m}_2(C)=0.01$.\\
\specialrule{0em}{6pt}{6pt}
\end{tabular}

By utilising Eq.~(\ref{eq_GDempsterrule1}) of the generalized Dempster's rule of combination, the fusing results are generated as follows:

\begin{tabular}[t]{l}
$\mathbf{m}(C)$ = 1. \\
\specialrule{0em}{6pt}{6pt}
\end{tabular}

Based on Eq.~(\ref{eq_Dempsterrule1}) of the classical Dempster's rule of combination, the fusing results are calculated as follows:

\begin{tabular}[t]{l}
$m(C)$ = 1. \\
\specialrule{0em}{6pt}{6pt}
\end{tabular}

In this example, $\mathbf{m}_1$ highly conflicts with $\mathbf{m}_2$, because $\mathbf{m}_1$ has a great belief value 0.99 on the object $A$, while $\mathbf{m}_2$ has a great belief value 0.99 on the object $B$.
However, as shown in the results, we can notice that when fusing the highly conflicting evidences, counter-intuitive results occur no matter we use the the generalized Dempster's rule of combination or the classical Dempster's rule of combination.

\begin{exmp}\label{exa_5}
\rm Supposing that there are two highly conflicting CBBAs $\mathbf{m}_1$ and $\mathbf{m}_2$ in the frame of discernment $\Omega=\{A,B,C\}$, and the two CBBAs are given as follows:
\end{exmp}

\begin{tabular}[t]{l}
$\mathbf{m}_1:$
$\mathbf{m}_1(A)=0.9900+0.1411i$,
$\mathbf{m}_1(C)=0.0100-0.1411i$;\\
$\mathbf{m}_2:$
$\mathbf{m}_2(B)=0.9900+0.1411i$,
$\mathbf{m}_2(C)=0.0100-0.1411i$.\\
\specialrule{0em}{6pt}{6pt}
\end{tabular}

The fusing results are calculated by utilising Eq.~(\ref{eq_GDempsterrule1}) of the generalized Dempster's rule of combination as follows:

\begin{tabular}[t]{l}
$\mathbf{m}(C)$ = 1.0000 + 0.0000i. \\
\specialrule{0em}{6pt}{6pt}
\end{tabular}

Through Example~\ref{exa_4} and Example~\ref{exa_5}, it is implied that counter-intuitive results occur when fusing the highly conflicting evidences modelled by either real numbers or complex numbers via the generalized Dempster's rule of combination.

\section{Application}\label{Application}
In this section, the proposed method is incorporated in quantum dynamical model, where the experimental data sets in~\cite{Busemeyer2009Empirical,wang2016interference} are used for the comparison with the related methods.

\subsection{Problem statement}
\begin{figure*}[htbp]
\centering
\subfigure[Narrow category]{
\label{examplefaces:a} 
\includegraphics[width=.7\linewidth,clip]{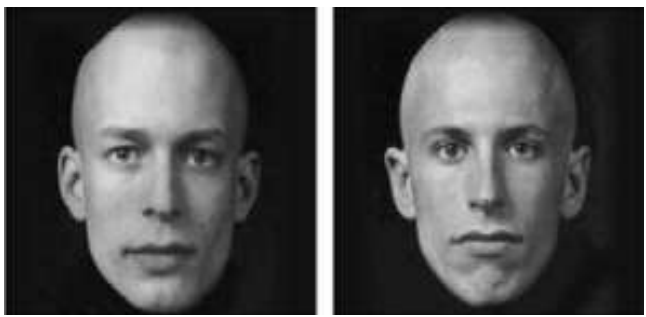}
}
\subfigure[Wide category]{
\label{examplefaces:b} 
\centering
\includegraphics[width=.7\linewidth,clip]{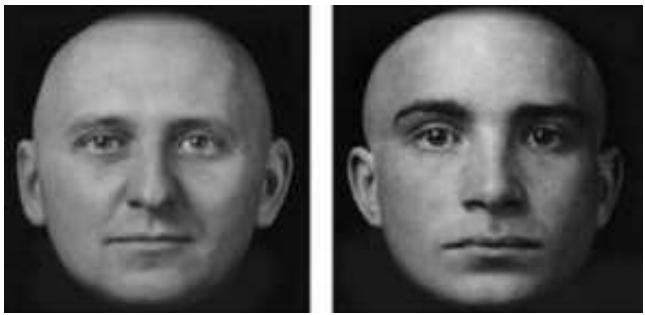}
}
\caption{Example faces used in a categorisation-decision experiment.}
\label{examplefaces} 
\end{figure*}

A new paradigm was presented by Townsend et al.~\cite{Townsend2000Exploring} in 2000 year to investigate the interactions between categorisation and decision-making.
Initially, this new paradigm was utilised to test a Markov model.
Afterward, it was extended for comparisons of Markov and quantum dynamical models by Busemeyer et al.~\cite{Busemeyer2009Empirical} in 2009.
In a categorisation (C) - decision (D) task, two different distributions of faces were utilised and shown to participants on each trial.
In particular, for a ``narrow'' face distribution, it had a narrow width and thick lips on average as shown in Fig.~\ref{examplefaces:a}; for a ``wide'' face distribution, it had a wide width and thin lips on average as shown in Fig.~\ref{examplefaces:b}.
The participants were requested to categorise the faces as a ``good'' guy or ``bad'' guy group, and/or they were requested to decide whether to take an ``attack'' or ``withdraw'' action.
The participants were notified that ``narrow'' faces had a 0.60 probability or chance to come from the ``bad'' guy population, while ``wide'' faces had a 0.60 probability or chance to come from the ``good'' guy population.
Thereinto, two test conditions, namely, a C-then-D condition and a D-alone condition were implemented to each participant across a series of trials.
Under the C-then-D condition, participants were requested to categorise the faces first, then made an action decision.
Differ with the C-then-D condition, participants only were requested to make an action decision without categorisation under the D-alone condition.
The experiment included 26 participants in total, of which for the C-then-D condition, each participant given 51 observations producing $26 \times 51$ = 1326 observations, and for the D-alone condition, each person provided 17 observations producing $17 \times 26$ = 442 total observations.

\begin{table*}[tp]\renewcommand{\arraystretch}{1.2}
\centering
{\footnotesize
\caption{Experimental results of a categorisation-decision task.}\label{categorydecisiontask}
\begin{tabular*}{\textwidth}{@{\extracolsep{\fill}}@{~~}llllllll@{~~}}
\toprule
Type face&$P(G)$&$P(A|G)$&$P(B)$&$P(A|B)$&$P_T(A)$&$P(A)$&$t$ \\\midrule
Wide  & 0.84 & 0.35 & 0.16 & 0.52 & 0.37 & 0.39 & 0.5733 \\
Narrow& 0.17 & 0.41 & 0.83 & 0.63 & 0.59 & 0.69 & 2.54 \\
\bottomrule
\end{tabular*}
}
\end{table*}

The experimental results were shown in Table~\ref{categorydecisiontask}.
The column labeled $P(G)$ denotes the probability of categorising the face as a ``good'' guy;
the column labeled $P(A|G)$ shows the probability of attacking when the face was categorised as a ``good'' guy.
The column labeled $P(B)$ represents the probability of categorising the face as a ``bad'' guy;
the column labeled $P(A|B)$ shows the probability of attacking when the face was categorised as a ``bad'' guy.
Then, the column labeled $P_T(A)$ represents the total probability of attacking as
\begin{equation}\label{eq_totalprobability}
P_T(A) = P(G) \cdot P(A|G) + P(B) \cdot P(A|B).
\end{equation}
On the other hand, the column labeled $P(A)$ denotes the probability of attacking when this decision was made alone.

In accordance with the law of total probability, the probability of attacking was supposed to be equal under two conditions.
Nevertheless, some deviation between $P_T(A)$ and $P(A)$ were generated for both faces as shown in Table~\ref{categorydecisiontask}.
Especially, for the narrow faces, the most pronounced deviation arose that caused a large positive interference effect.
Through a paired $t$-test to measure the significance of the difference between $P_T(A)$ and $P(A)$, the results indicated that the mean interference effect was statistically significant for the narrow faces, but not for the wide faces.
In this study, therefore, the interference effect is investigated and analysed in terms of attacking actions towards the narrow faces.

\subsection{Implementation}
In this section, the proposed method is integrated into quantum dynamical model to model the human decision making process in an evidential framework.

\subsubsection{Representation of beliefs and actions}
\begin{figure}[!htpb]
\centering
\includegraphics[width=9cm,clip]{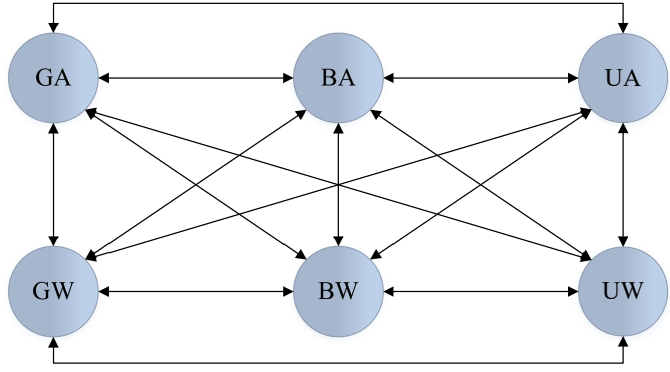}
 \caption{Transition diagram in an evidential quantum dynamical model.}\label{transition}
\end{figure}

In an evidential quantum dynamical model, the categorisation (C) - decision (D) experiment involves a set of six exhaustive outcomes $\{C_GD_A$, $C_GD_W$, $C_BD_A$, $C_BD_W$, $C_UD_A$, $C_UD_W\}$, where, for instance, $C_GD_A$ symbolises the event in which the participant believes the face as a ``good'' (G) guy, but the participant intends to take an ``attack'' (A) action, while $C_UD_W$ symbolises the event in which the participant is skeptical or hesitating of the face as a ``good'' or ``bad'' (B) guy that is in a an uncertain (U) condition, but the participant intends to act by withdrawing (W).
The evidential quantum dynamical model assumes that these six events correspond to six basis belief-action states of the decision maker $\{|C_GD_A\rangle$, $|C_GD_W\rangle$, $|C_BD_A\rangle$, $|C_BD_W\rangle$, $|C_UD_A\rangle$, $|C_UD_W\rangle\}$.
All the possible transitions between the six basis states in an evidential quantum dynamical model are depicted in Fig.~\ref{transition}.

At the beginning of a categorisation-decision task, the participant has some possibilities to be in every basis state in Fig.~\ref{transition}.
Hence, the state of a participant is a superposition of the six orthonormal basis states, denoted by
\begin{equation}\label{eq_superposition}
\begin{aligned}
|\psi\rangle =
&\psi_{GA} \cdot |C_GD_A\rangle + \psi_{GW} \cdot |C_GD_W\rangle + \psi_{BA} \cdot |C_BD_A\rangle + \psi_{BW} \cdot |C_BD_W\rangle \\
&+ \psi_{UA} \cdot |C_UD_A\rangle + \psi_{UW} \cdot |C_UD_W\rangle.
\end{aligned}
\end{equation}

An amplitude distribution corresponding to the initial state is denoted by the following $6 \times 1$ column matrix,
\begin{equation}
{
\psi(0)=
\begin{bmatrix}
\psi_{GA} \\
\psi_{GW} \\
\psi_{BA} \\
\psi_{BW} \\
\psi_{UA} \\
\psi_{UW} \\
\end{bmatrix},\quad
}\end{equation}
where $|\psi_{ij}|$ represents the probability of observing basis state $|C_iD_j\rangle$ initially in which $i \in \{G, B, U\}$ and $j \in \{A, W\}$.
The squared length of $\psi$ must be equal to one, such that $\psi^\dag \cdot \psi = 1$, where $\psi^\dag$ is the conjugate of $\psi$.
Here, the probability of initial state is assumed to be distributed averagely.

\subsubsection{Inferences based on prior information}
In the course of decision making process, the initial state $\psi(0)$ with regard to the participant's beliefs at time $t=0$ is turned into a new state $\psi(t_1)$ at time $t_1$.
For the evidential quantum dynamical model, the categorisation of faces are decided by participants under the C-then-D condition.
When the face is classified as a ``good'' guy, the amplitude distribution across the basis states becomes
\begin{equation}
{
\psi(t_1)= \frac{1}{\sqrt{|\psi_{GA}|^2+|\psi_{GW}|^2}}
\begin{bmatrix}
\psi_{GA} \\
\psi_{GW} \\
0 \\
0 \\
0 \\
0 \\
\end{bmatrix}
=
\begin{bmatrix}
\psi_{G} \\
\mathbf{0} \\
\mathbf{0} \\
\end{bmatrix},\quad
}\end{equation}
in which $\sqrt{|\psi_{GA}|^2+|\psi_{GW}|^2}$ represents the initial probability of categorising the face as a ``good'' guy.
This $2 \times 1$ matrix $\psi_{G}$ has a squared length that is equal to one.
It is a conditional amplitude distribution across actions under the situation where the face is classified as a ``good'' guy.

When the face is categorised as a ``bad'' guy, the amplitude distribution across the basis states states turns into
\begin{equation}
{
\psi(t_1)= \frac{1}{\sqrt{|\psi_{BA}|^2+|\psi_{BW}|^2}}
\begin{bmatrix}
0 \\
0 \\
\psi_{BA} \\
\psi_{BW} \\
0 \\
0 \\
\end{bmatrix}
=
\begin{bmatrix}
\mathbf{0} \\
\psi_{B} \\
\mathbf{0} \\
\end{bmatrix},\quad
}\end{equation}
in which $\sqrt{|\psi_{BA}|^2+|\psi_{BW}|^2}$ is the initial probability of categorising the face as a ``bad'' guy.
This $2 \times 1$ matrix $\psi_{B}$ has a squared length that is equal to one.
It is a conditional amplitude distribution across actions under the situation where the face is classified as a ``bad'' guy.

When the face cannot be categorised as a ``good'' or ``bad'' guy due to the skepticism or hesitation of participant, the amplitude distribution across the basis states becomes
\begin{equation}
{
\psi(t_1)= \frac{1}{\sqrt{|\psi_{UA}|^2+|\psi_{UW}|^2}}
\begin{bmatrix}
0 \\
0 \\
0 \\
0 \\
\psi_{UA} \\
\psi_{UW} \\
\end{bmatrix}
=
\begin{bmatrix}
\mathbf{0} \\
\mathbf{0} \\
\psi_{U} \\
\end{bmatrix},\quad
}\end{equation}
in which $\sqrt{|\psi_{UA}|^2+|\psi_{UW}|^2}$ denotes the initial probability that the participant cannot categorise the face as a ``good'' or ``bad'' guy because of lacking sufficient information.
This $2 \times 1$ matrix $\psi_{U}$ has a squared length that is equal to one.
It is a conditional amplitude distribution across actions under the case where the face cannot be classified and it is in an uncertain situation.

Under the D alone condition, because the participant is not requested to categorise the faces before taking an action, there is no new information involved in terms of categorisation.
Therefore, the amplitude distribution across the basis states remains the same as the initial one
\begin{equation}\label{eq_aloneinitialstate}
\begin{aligned}
\psi(t_1)= \psi(0)=&
\begin{bmatrix}
\sqrt{|\psi_{GA}|^2+|\psi_{GW}|^2} \cdot \psi_{G} \\
\sqrt{|\psi_{BA}|^2+|\psi_{BW}|^2} \cdot \psi_{B} \\
\end{bmatrix} \\
&=
\sqrt{|\psi_{GA}|^2+|\psi_{GW}|^2}
\begin{bmatrix}
\psi_{G} \\
\mathbf{0} \\
\end{bmatrix} 
+
\sqrt{|\psi_{BA}|^2+|\psi_{BW}|^2}
\begin{bmatrix}
\mathbf{0} \\
\psi_{B} \\
\end{bmatrix},
\end{aligned}
\end{equation}
where it represents the initial state under a condition without categorisation as a superposition which is a weighted sum of the amplitude distributions for the two conditions.

\subsubsection{Strategies based on payoffs}
In order to choose an appropriate action, a decision maker needs to assess the payoffs, so that it turns the previous state $\psi(t_1)$ at time $t_1$ into a new state $\psi(t_2)$ at time $t_2$.
The state evolution during this time period $t_2 - t_1$ corresponds to the thought process resulting in a decision.
For the evidential quantum dynamical model, the evolution of the state obeys a Schr$\ddot{o}$dinger equation during the decision making process which is driven by a $6 \times 6$ Hamiltonian matrix $H$:
\begin{equation}
\frac{d}{dt}\psi(t) = -i \cdot H \cdot \psi(t),
\end{equation}
where $H$ is a Hermitian matrix: $H^\dag = H$ that will be discussed below.

It has the following matrix exponential solution for $t = t_2 - t_1$,
\begin{equation}
\psi(t_2) = e^{-iHt} \cdot \psi(t_1),
\end{equation}
where $\psi(t_2)$ represents the amplitude distribution across states after evolution by evaluating the payoffs, and a unitary matrix is defined by
\begin{equation}
U(t) = e^{-iHt},
\end{equation}
which determines the transition probabilities.

Here, the Hamiltonian matrix $H$ is defined by
\begin{equation}
H=
\begin{bmatrix}
H_G & \mathbf{0} & \mathbf{0} \\
\mathbf{0} & H_B & \mathbf{0} \\
\mathbf{0} & \mathbf{0} & H_U \\
\end{bmatrix},
\end{equation}
where
\begin{equation}
\begin{aligned}
H_G= \frac{1}{1+h_G^2}
\begin{bmatrix}
h_G & 1 \\
1 & -h_G \\
\end{bmatrix}, \\
H_B= \frac{1}{1+h_B^2}
\begin{bmatrix}
h_B & 1 \\
1 & -h_B \\
\end{bmatrix}, \\
H_U= \frac{1}{1+h_U^2}
\begin{bmatrix}
h_U & 1 \\
1 & -h_U \\
\end{bmatrix}.
\end{aligned}
\end{equation}

When the face is categorised as a ``good'' guy by the participant, the $2 \times 2$ Hamiltonian matrix $H_G$ is supposed to be utilised, while when the face is categorised as a ``bad'' guy by the participant, the $2 \times 2$ Hamiltonian matrix $H_B$ should be used.
If the participant cannot categorise the face as a ``good'' or ``bad'' guy which is in an uncertain state, the $2 \times 2$ Hamiltonian matrix $H_U$ will be applied.
To be specific, the parameter $h_G$ is a function of the difference between the payoffs for attacking with respect to withdrawing when categorising the face as a ``good'' guy;
the parameter $h_B$ is a function of the difference between the payoffs for
attacking with respect to withdrawing when categorising the face as a ``bad'' guy;
the parameter $h_U$ is a function of the difference between the payoffs for
attacking with respect to withdrawing when the participant cannot categorise the face.
The Hamiltonian matrix transforms the state probabilities to favor either attacking or withdrawing according to the payoff in terms of each belief state.

Afterwards, the state of the participant at time $t_2$ can be obtained.
In the C-then-D condition, when the face is classified as a ``good'' guy, the state $\psi(t_1)$ at time $t_1$ changes into the state $\psi(t_2)$ at time $t_2$ by
\begin{equation}
\psi(t_2)= e^{-iHt} \cdot \psi(t_1)=
\begin{bmatrix}
e^{-iH_Gt} & \mathbf{0} & \mathbf{0} \\
\mathbf{0} & e^{-iH_Bt} & \mathbf{0} \\
\mathbf{0} & \mathbf{0} & e^{-iH_Ut} \\
\end{bmatrix}
\cdot 
\begin{bmatrix}
\psi_{G} \\
\mathbf{0} \\
\mathbf{0} \\
\end{bmatrix}
= e^{-iH_Gt} \cdot \psi_G.
\end{equation}

When the face is classified as a ``bad'' guy, the state $\psi(t_1)$ at time $t_1$ turns into the state $\psi(t_2)$ at time $t_2$ by
\begin{equation}
\psi(t_2)= e^{-iHt} \cdot \psi(t_1)=
\begin{bmatrix}
e^{-iH_Gt} & \mathbf{0} & \mathbf{0} \\
\mathbf{0} & e^{-iH_Bt} & \mathbf{0} \\
\mathbf{0} & \mathbf{0} & e^{-iH_Ut} \\
\end{bmatrix}
\cdot 
\begin{bmatrix}
\mathbf{0} \\
\psi_{B} \\
\mathbf{0} \\
\end{bmatrix}
= e^{-iH_Bt} \cdot \psi_G.
\end{equation}

When the participant cannot categorise the face as a ``good'' or ``bad'' guy, the state $\psi(t_1)$ at time $t_1$ becomes the state $\psi(t_2)$ at time $t_2$ by
\begin{equation}
\psi(t_2)= e^{-iHt} \cdot \psi(t_1)=
\begin{bmatrix}
e^{-iH_Gt} & \mathbf{0} & \mathbf{0} \\
\mathbf{0} & e^{-iH_Bt} & \mathbf{0} \\
\mathbf{0} & \mathbf{0} & e^{-iH_Ut} \\
\end{bmatrix}
\cdot 
\begin{bmatrix}
\mathbf{0} \\
\mathbf{0} \\
\psi_{U} \\
\end{bmatrix}
= e^{-iH_Ut} \cdot \psi_G.
\end{equation}

On the other hand, in the D alone condition, the state $\psi(0)$ at time $t=0$ turns into the state $\psi(t_2)$ at time $t_2$ by
\begin{equation}\label{eq_aloneinitialstate}
\begin{aligned}
\psi(t_2)=& e^{-iHt} \cdot \psi(0)=
\begin{bmatrix}
e^{-iH_Gt} & \mathbf{0} \\
\mathbf{0} & e^{-iH_Bt} \\
\end{bmatrix}
\cdot 
\begin{bmatrix}
\sqrt{|\psi_{GA}|^2+|\psi_{GW}|^2} \cdot \psi_{G} \\
\sqrt{|\psi_{BA}|^2+|\psi_{BW}|^2} \cdot \psi_{B} \\
\end{bmatrix} \\
=
&\sqrt{|\psi_{GA}|^2+|\psi_{GW}|^2} \cdot e^{-iH_Gt} \cdot \psi_{G} + \sqrt{|\psi_{BA}|^2+|\psi_{BW}|^2} \cdot e^{-iH_Bt} \cdot \psi_{B},
\end{aligned}
\end{equation}
where it expresses the state $\psi(t_2)$ at time $t_2$ under unknown categorisation condition as a superposition which is a weighted sum of the amplitude distributions for the two cases.

\subsubsection{Predictions of the evidential quantum dynamical model}
In the evidential quantum dynamical model, the interference effect can be predicted based on the state evolution of the participant.
In order to predict a state of attacking with regard to a certain categorisation of face, a measure matrix $M$ is defined by
\begin{equation}
M=
\begin{bmatrix}
M_{G} & \mathbf{0} & \mathbf{0} \\
\mathbf{0} & M_{B} & \mathbf{0} \\
\mathbf{0} & \mathbf{0} & M_{U} \\
\end{bmatrix},
\end{equation}
where when the face is categorised as a ``good'' guy by the participant, the $2 \times 2$ measure matrix $M_{G}$ is supposed to be utilised;
when the face is categorised as a ``bad'' guy by the participant, the $2 \times 2$ Hamiltonian matrix $M_{B}$ should be used;
if the participant cannot categorise the face as a ``good'' or ``bad'' guy, the $2 \times 2$ Hamiltonian matrix $M_{U}$ will be applied.

In the C-then-D condition, for measuring the belief of attacking with respect to the situation where the face is categorised as a ``good'' guy, the $2 \times 2$ measure matrices $M_{G}$, $M_{B}$ and $M_{U}$ are set as follows
\begin{equation}
\begin{aligned}
M_{G} =
\begin{bmatrix}
1 & 0 \\
0 & 0 \\
\end{bmatrix}, \quad
M_{B} =
\begin{bmatrix}
1 & 0 \\
0 & 0 \\
\end{bmatrix}, \quad
M_{U} =
\begin{bmatrix}
1 & 0 \\
0 & 0 \\
\end{bmatrix}.
\end{aligned}
\end{equation}

Then, the prediction of the attacking belief for the three cases, i.e., $\Phi(A|G)$, $\Phi(A|B)$ and $\Phi(A|U)$ can be obtained by the following equations, respectively
\begin{equation}
\begin{aligned}
\Phi(A|G) &=  M \cdot e^{-iHt} \cdot \psi(t_1) \\
&=
\begin{bmatrix}
M_{G} & \mathbf{0} & \mathbf{0} \\
\mathbf{0} & M_{B} & \mathbf{0} \\
\mathbf{0} & \mathbf{0} & M_{U} \\
\end{bmatrix} 
\cdot
\begin{bmatrix}
e^{-iH_Gt} & \mathbf{0} & \mathbf{0} \\
\mathbf{0} & e^{-iH_Bt} & \mathbf{0} \\
\mathbf{0} & \mathbf{0} & e^{-iH_Ut} \\
\end{bmatrix} 
\cdot
\begin{bmatrix}
\psi_G \\
\mathbf{0} \\
\mathbf{0} \\
\end{bmatrix} \\
&=
M_{G} \cdot e^{-iH_Gt} \cdot \psi_G.
\end{aligned}
\end{equation}

\begin{equation}
\begin{aligned}
\Phi(A|B) &=  M \cdot e^{-iHt} \cdot \psi(t_1) \\
&=
\begin{bmatrix}
M_{G} & \mathbf{0} & \mathbf{0} \\
\mathbf{0} & M_{B} & \mathbf{0} \\
\mathbf{0} & \mathbf{0} & M_{U} \\
\end{bmatrix} 
\cdot
\begin{bmatrix}
e^{-iH_Gt} & \mathbf{0} & \mathbf{0} \\
\mathbf{0} & e^{-iH_Bt} & \mathbf{0} \\
\mathbf{0} & \mathbf{0} & e^{-iH_Ut} \\
\end{bmatrix} 
\cdot
\begin{bmatrix}
\mathbf{0} \\
\psi_B \\
\mathbf{0} \\
\end{bmatrix} \\
&=
M_{B} \cdot e^{-iH_Bt} \cdot \psi_B.
\end{aligned}
\end{equation}

\begin{equation}
\begin{aligned}
\Phi(A|U) &=  M \cdot e^{-iHt} \cdot \psi(t_1) \\
&=
\begin{bmatrix}
M_{G} & \mathbf{0} & \mathbf{0} \\
\mathbf{0} & M_{B} & \mathbf{0} \\
\mathbf{0} & \mathbf{0} & M_{U} \\
\end{bmatrix} 
\cdot
\begin{bmatrix}
e^{-iH_Gt} & \mathbf{0} & \mathbf{0} \\
\mathbf{0} & e^{-iH_Bt} & \mathbf{0} \\
\mathbf{0} & \mathbf{0} & e^{-iH_Ut} \\
\end{bmatrix} 
\cdot
\begin{bmatrix}
\mathbf{0} \\
\mathbf{0} \\
\psi_U \\
\end{bmatrix} \\
&=
M_{U} \cdot e^{-iH_Ut} \cdot \psi_U.
\end{aligned}
\end{equation}

Then, on the basis of Eq.~(\ref{eq_pignistic}), the belief of uncertain case that the face cannot be classified by the participant will be assigned to another two certain cases equally, denoted by
\begin{eqnarray}
\Phi'(A|G) &=  \Phi(A|G) + \frac{1}{2} \Phi(A|U), \\
\Phi'(A|B) &=  \Phi(A|B) + \frac{1}{2} \Phi(A|U),
\end{eqnarray}
where $\Phi'(A|G)$ represents the conditional amplitude of attacking when the face is categorised as a ``good'' guy with the involvement of uncertain information, which has a squared length equal to $P(A|G)$;
$\Phi'(A|B)$ represents the conditional amplitude of attacking when the face is categorised as a ``bad'' guy involving the uncertain information, which has a squared length equal to $P(A|B)$.

After that, the prediction of total probability for attacking under the C-then-D condition can be calculated by
\begin{equation}
\begin{aligned}
P(A) = &P(G) \cdot P(A|G) + P(B) \cdot P(A|B) \\
= &(|\psi_{GA}|^2+|\psi_{GW}|^2) \cdot \| \Phi'(A|G) \|^2 +
(|\psi_{BA}|^2+|\psi_{BW}|^2) \cdot \| \Phi'(A|B) \|^2 \\
= &(|\psi_{GA}|^2+|\psi_{GW}|^2) \cdot \| (M_{G}+\frac{1}{2}M_{U}) \cdot e^{-iH_Gt} \cdot \psi_G \|^2 + \\
&(|\psi_{BA}|^2+|\psi_{BW}|^2) \cdot \| (M_{B}+\frac{1}{2}M_{U}) \cdot e^{-iH_Bt} \cdot \psi_B \|^2.
\end{aligned}
\end{equation}

In the D alone condition, in order to measure the belief of attacking without categorisation, the $2 \times 2$ measure matrices $M_{GA}$ and $M_{BA}$ are set as follows
\begin{equation}
\begin{aligned}
M_{G} =
\begin{bmatrix}
1 & 0 \\
0 & 0 \\
\end{bmatrix}, \quad
M_{B} =
\begin{bmatrix}
0 & 0 \\
0 & 1 \\
\end{bmatrix}. \quad
\end{aligned}
\end{equation}

Thus, the prediction of total probability for attacking without categorisation can be computed by
\begin{equation}
\begin{aligned}
P(A) = &\| M \cdot \psi(t_2) \|^2 = \| M \cdot e^{-iHt} \cdot \psi(0) \|^2 \\
= &\left\|
\begin{bmatrix}
M_{G} & \mathbf{0} \\
\mathbf{0} & M_{B} \\
\end{bmatrix} 
\cdot
\begin{bmatrix}
e^{-iH_Gt} & \mathbf{0} \\
\mathbf{0} & e^{-iH_Bt} \\
\end{bmatrix} 
\cdot
\begin{bmatrix}
\sqrt{|\psi_{GA}|^2+|\psi_{GW}|^2} \cdot \psi_{G} \\
\sqrt{|\psi_{BA}|^2+|\psi_{BW}|^2} \cdot \psi_{B} \\
\end{bmatrix} 
\right\|^2 \\
= &\|
(|\psi_{GA}|^2+|\psi_{GW}|^2) \cdot M_G \cdot e^{-iH_Gt} \cdot \psi_G +
(|\psi_{BA}|^2+|\psi_{BW}|^2) \cdot M_B \cdot e^{-iH_Bt} \cdot \psi_B 
\|^2.
\end{aligned}
\end{equation}

We can notice that the prediction of total probability for attacking under the C-then-D condition is different comparing with the D alone condition.
The difference of probability between these two conditions indicates the interference effect caused by the interactions between categorisation and decision-making.
Specifically, the belief of uncertain state is modelled and transferred into another two certain states under the C-then-D condition.
Whereas, the function $\Phi(A|U)$ is not generated under the D alone condition, since the action of attacking is taken without categorisation.
As a results, the interference effect resulted from the interactions between categorisation and decision-making can be predicted under these two different conditions.

\subsection{Experimental results}
\subsubsection{Parameter setting}
In the experiments, based on literatures~\cite{He2017Anevidential,busemeyer2012quantum,Busemeyer2009Empirical,wang2016interference}, on account of realising the predictions for the evidential quantum dynamical model, the time process parameter $t$ is set as $\frac{\pi}{2}$ to allow the selection probability to achieve maximum across time.
$P(G)$ or $P(B)$ is set as the same with the relevant observed experimental results.
Meanwhile, three free parameters, $h_G$, $h_B$ and $h_U$ are estimated under the C-then-D condition, while two free parameters, $h_G$ and $h_B$ are estimated under the D alone condition.
These free parameters are fitted by minimising the sum of squared errors (SSE) between the predicted and observed mean probability judgments for each of the two conditions.

\subsubsection{Comparisons of different models}
In order to validate the feasibility and effectiveness of the proposed evidential quantum dynamical model, it is compared with the Markov belief-action (MBA) model~\cite{Townsend2000Exploring}, the quantum belief-action entanglement (QBAE) model~\cite{Busemeyer2009Empirical}, and the evidential Markov (EM) model~\cite{He2017Anevidential}.
The comparisons of prediction results for the categorisation-decision task under two above-mentioned conditions (i.e., the C-then-D condition and the D alone condition) are shown in Table~\ref{comparison}, in which $Obs$ denotes the observed experimental results from the literatures~\cite{Busemeyer2009Empirical,wang2016interference}.

The columns labeled $P(A|G)$ and $P(A|B)$ represent the predicted probabilities of attacking when the faces are classified as a ``good'' guy and a ``bad'' guy under the C-then-D condition, respectively.
The column labeled $P_T(A)$ denotes the predicted total probability of attacking under the C-then-D condition, while the column labeled $P(A)$ represents the predicted total probability of attacking under the D alone condition.

\begin{sidewaystable}[!htbp]\renewcommand{\arraystretch}{1.2}
{\footnotesize
\caption{Comparisons of prediction in terms of different models.}\label{comparison}
\begin{tabular*}{\textwidth}{@{\extracolsep{\fill}}@{~~}llllllll@{~~}}
\toprule
Literature       &Method&$P(G)$&$P(A|G)$&$P(B)$&$P(A|B)$&$P_T$&$P(A)$ \\\midrule
Busemeyer et al.~\cite{Busemeyer2009Empirical}
& $Obs$            &0.17&0.41&0.83&0.63&0.59&0.69  \\
& EM             &0.17&0.39&0.83&0.61&0.57&0.69  \\
& QBAE           &0.17&0.41&0.83&0.66&0.62&0.68  \\
& MBA            &0.17&0.40&0.83&0.63&0.59&0.59  \\
& Proposed method&0.17&0.41&0.83&0.56&0.53&0.60 \\\midrule
Wang \& Busemeyer
& $Obs$            &0.21&0.41&0.79&0.58&0.54&0.59  \\
Experiment 1~\cite{wang2016interference}
& EM             &0.21&0.42&0.79&0.58&0.55&0.60  \\
& QBAE           &0.21&0.45&0.79&0.54&0.52&0.57  \\
& MBA            &0.21&0.39&0.79&0.60&0.55&0.55  \\
& Proposed method&0.21&0.41&0.79&0.56&0.53&0.58  \\\midrule
Wang \& Busemeyer
& $Obs$            &0.24&0.37&0.76&0.61&0.55&0.60  \\
Experiment 2~\cite{wang2016interference}
& EM             &0.24&0.38&0.76&0.62&0.56&0.61  \\
& QBAE           &0.21&0.33&0.79&0.68&0.61&0.63  \\
& MBA            &0.23&0.39&0.77&0.66&0.60&0.59  \\
& Proposed method&0.24&0.41&0.76&0.56&0.52&0.56  \\\midrule
Wang \& Busemeyer
& $Obs$            &0.24&0.33&0.76&0.66&0.58&0.62  \\
Experiment 3~\cite{wang2016interference}
& EM             &0.25&0.34&0.75&0.66&0.58&0.64  \\
& QBAE           &0.21&0.32&0.79&0.69&0.61&0.63  \\
& MBA            &0.23&0.47&0.77&0.55&0.53&0.53  \\
& Proposed method&0.24&0.35&0.76&0.56&0.51&0.55  \\\midrule
Average
& $Obs$            &0.22&0.38&0.79&0.62&0.57&0.63  \\
& EM             &0.22&0.38&0.78&0.62&0.57&0.64  \\
& QBAE           &0.20&0.38&0.80&0.64&0.59&0.63  \\
& MBA            &0.21&0.41&0.79&0.61&0.57&0.57  \\
& Proposed method&0.22&0.39&0.79&0.56&0.52&0.57  \\
\bottomrule
\end{tabular*}
}
\end{sidewaystable}

As shown in Table~\ref{comparison}, it is obvious that the predicted results of the proposed model, namely, $P(G)$, $P(A|G)$, $P(B)$, $P(A|B)$, $P_T$ and $P(A)$ in terms of different data sets are very close to the observed results.
On the other hand, the prediction of total probability for attacking under the C-then-D condition, i.e., $P_T$ is different comparing with the D alone condition, i.e., $P(A)$ where the deviation between $P_T$ and $P(A)$ indicates the predicted interference effect caused in the categorisation-decision task.
Hence, it can be concluded that the proposed evidential quantum dynamical model is as feasible and effectiveness as the related QBAE model~\cite{Busemeyer2009Empirical} and EM model~\cite{He2017Anevidential}.
Whereas, the MBA model~\cite{Townsend2000Exploring} cannot predict the interference effect due to following the law of total probability.

\section{Conclusions}\label{Conclusion}
Dempster--Shafer evidence theory is a very useful uncertainty reasoning tool in modeling and handling uncertainties regardless of prior information.
On the other hand, the quantum theory has proven its powerful capabilities of solving the decision making problems.
Whereas, the classical Dempster--Shafer evidence theory expressed by real numbers can not be integrated directly with the quantum theory.
So, how can we establish a bridge of communications between the classical Dempster--Shafer evidence theory and the quantum theory?
To address this issue, in this study, a generalized Dempster--Shafer evidence theory which is expressed by complex numbers is proposed.
Unlike the existing evidence theory, a mass function in the generalized Dempster--Shafer evidence theory is modelled as a complex number, called as a complex mass function.
When the complex mass function is degenerated from complex numbers to real numbers, the generalized Dempster--Shafer evidence theory degenerates to the classical evidence theory under the condition that the conflict coefficient between the evidences $\mathbf{K}$ is less than 1.
Finally, an application of an evidential quantum dynamical model is implemented by integrating the generalized Dempster--Shafer evidence theory with the quantum dynamical model.
The experimental results validate the feasibility and effectiveness of the proposed method.

\section*{Competing interests}
The author declare that she have no competing interests.

\section*{Acknowledgments}
This research is supported by the Fundamental Research Funds for the Central Universities (Grant No. SWU115008), the National Natural Science Foundation of China (Nos. 61672435, 61702427, 61702426) and the 1000-Plan of Chongqing by Southwest University (No. SWU116007).

\normalsize

\begin{thebibliography}{10}
\expandafter\ifx\csname url\endcsname\relax
  \def\url#1{\texttt{#1}}\fi
\expandafter\ifx\csname urlprefix\endcsname\relax\def\urlprefix{URL }\fi
\expandafter\ifx\csname href\endcsname\relax
  \def\href#1#2{#2} \def\path#1{#1}\fi

\bibitem{walczak1999rough}
B.~Walczak, D.~Massart, Rough sets theory, Chemometrics and Intelligent
  Laboratory Systems 47~(1) (1999) 1--16.

\bibitem{zadeh1965fuzzy}
L.~A. Zadeh, Fuzzy sets, Information and Control 8~(3) (1965) 338--353.

\bibitem{liu2013fuzzy}
H.-C. Liu, L.~Liu, Q.-L. Lin, Fuzzy failure mode and effects analysis using
  fuzzy evidential reasoning and belief rule-based methodology, IEEE
  Transactions on Reliability 62~(1) (2013) 23--36.

\bibitem{Den2011Maximum}
T.~Den{\oe}ux, Maximum likelihood estimation from fuzzy data using the {EM}
  algorithm, Fuzzy Sets \& Systems 183~(1) (2011) 72--91.

\bibitem{zhengrong2017ADAC}
R.~Zhang, B.~Ashuri, Y.~Deng, A novel method for forecasting time series based
  on fuzzy logic and visibility graph, Advances in Data Analysis and
  Classification 11 (2017) 759--783.

\bibitem{Dempster1967Upper}
A.~P. Dempster, Upper and lower probabilities induced by a multivalued mapping,
  Annals of Mathematical Statistics 38~(2) (1967) 325--339.

\bibitem{Jiang2017mGCR}
W.~Jiang, J.~Zhan, A modified combination rule in generalized evidence theory,
  Applied Intelligence 46~(3) (2017) 630--640.

\bibitem{xuhonghui2018}
H.~Xu, Y.~Deng, Dependent evidence combination based on {Shearman} coefficient
  and {Pearson} coefficient, IEEE Access 6~(1) (2018) 11634--11640.

\bibitem{XDWJIJIS21929}
X.~Deng, W.~Jiang, An evidential axiomatic design approach for decision making
  using the evaluation of belief structure satisfaction to uncertain target
  values, International Journal of Intelligent Systems 33~(1) (2018) 15--32.

\bibitem{zadeh2011note}
L.~A. Zadeh, A note on {Z-numbers}, Information Sciences 181~(14) (2011)
  2923--2932.

\bibitem{Kang2017Stable}
B.~Kang, G.~Chhipi-Shrestha, Y.~Deng, K.~Hewage, R.~Sadiq, Stable strategies
  analysis based on the utility of {Z-number} in the evolutionary games,
  Applied Mathematics and Computation 324 (2018) 202--217.

\bibitem{zhangqi2017}
Q.~Zhang, M.~Li, Y.~Deng, Measure the structure similarity of nodes in complex
  networks based on relative entropy, Physica A: Statistical Mechanics and its
  Applications 491 (2018) 749--763.

\bibitem{xiao2016intelligent}
F.~Xiao, An intelligent complex event processing with {D} numbers under fuzzy
  environment, Mathematical Problems in Engineering 2016 (2016) 1--10.

\bibitem{DAHPcredibility2018}
X.~Deng, Y.~Deng, D-{AHP} method with different credibility of information,
  Soft Computing (2018) DOI: 10.1007/s00500--017--2993--9.

\bibitem{liubaoyuIJCCC2017}
B.~Liu, Y.~Hu, Y.~Deng, New failure mode and effects analysis based on {D}
  numbers downscaling method, International Journal of Computers Communications
  \& Control 13~(2) (2018) In press.

\bibitem{Bian2018Failure}
T.~Bian, H.~Zheng, L.~Yin, Y.~Deng, Failure mode and effects analysis based on
  {D} numbers and {TOPSIS}, Quality and Reliability Engineering International
  (2018) DOI: 10.1002/qre.2268\href {http://dx.doi.org/10.1002/qre.2268}
  {\path{doi:10.1002/qre.2268}}.

\bibitem{zhou2017dependence}
X.~Zhou, X.~Deng, Y.~Deng, S.~Mahadevan, Dependence assessment in human
  reliability analysis based on {D} numbers and {AHP}, Nuclear Engineering and
  Design 313 (2017) 243--252.

\bibitem{yang2012belief}
J.-B. Yang, Y.-M. Wang, D.-L. Xu, K.-S. Chin, L.~Chatton, Belief rule-based
  methodology for mapping consumer preferences and setting product targets,
  Expert Systems with Applications 39~(5) (2012) 4749--4759.

\bibitem{fu2015group}
C.~Fu, J.-B. Yang, S.-L. Yang, A group evidential reasoning approach based on
  expert reliability, European Journal of Operational Research 246~(3) (2015)
  886--893.

\bibitem{yang2014interactive}
J.-B. Yang, D.-L. Xu, Interactive minimax optimisation for integrated
  performance analysis and resource planning, Computers \& Operations Research
  46 (2014) 78--90.

\bibitem{yang2013evidential}
J.-B. Yang, D.-L. Xu, Evidential reasoning rule for evidence combination,
  Artificial Intelligence 205 (2013) 1--29.

\bibitem{ma2012qualitative}
J.~Ma, Qualitative approach to {Bayesian} networks with multiple causes, IEEE
  Transactions on Systems, Man, and Cybernetics-Part A: Systems and Humans
  42~(2) (2012) 382--391.

\bibitem{Cuzzolin2014Learning}
F.~Cuzzolin, M.~Sapienza, Learning pullback {HMM} distances, IEEE Transactions
  on Pattern Analysis \& Machine Intelligence 36~(7) (2014) 1483--1489.

\bibitem{shafer1976mathematical}
G.~Shafer, et~al., A mathematical theory of evidence, Vol.~1, Princeton
  University Press Princeton, 1976.

\bibitem{jiang2018IJSS}
W.~Jiang, B.~Wei, X.~Liu, X.~Li, H.~Zheng, {Intuitionistic fuzzy evidential
  power aggregation operator and its application in multiple criteria
  decision-making}, International Journal of Systems Science 49 (2018)
  582--594.

\bibitem{feiliguo2017new}
L.~Fei, H.~Wang, L.~Chen, Y.~Deng, A new vector valued similarity measure for
  intuitionistic fuzzy sets based on {OWA} operators, Iranian Journal of Fuzzy
  Systems (2017) accepted.

\bibitem{jiang2017Intuitionistic}
W.~Jiang, B.~Wei, X.~Liu, X.~Li, H.~Zheng, Intuitionistic fuzzy power
  aggregation operator based on entropy and its application in decision making,
  International Journal of Intelligent Systems 33~(1) (2018) 49--67.

\bibitem{jiang2017evidence}
W.~Jiang, S.~Wang, X.~Liu, H.~Zheng, B.~Wei, Evidence conflict measure based on
  {OWA} operator in open world, PloS one 12~(5) (2017) e0177828.

\bibitem{deng2016evidenceIEEE}
X.~Deng, D.~Han, J.~Dezert, Y.~Deng, Y.~Shyr, Evidence combination from an
  evolutionary game theory perspective, IEEE Transactions on Cybernetics 46~(9)
  (2016) 2070--2082.

\bibitem{Wang2017IJCCC}
W.~Jiang, S.~Wang, An uncertainty measure for interval-valued evidences,
  International Journal of Computers Communications \& Control 12~(5) (2017)
  631--644.

\bibitem{denoeux1995k}
T.~Den{\oe}ux, A k-nearest neighbor classification rule based on
  {Dempster--Shafer} theory, IEEE Transactions on Systems, Man, and Cybernetics
  25~(5) (1995) 804--813.

\bibitem{ma2016evidential}
J.~Ma, W.~Liu, P.~Miller, H.~Zhou, An evidential fusion approach for gender
  profiling, Information Sciences 333 (2016) 10--20.

\bibitem{liu2016adaptive}
Z.-g. Liu, Q.~Pan, J.~Dezert, A.~Martin, Adaptive imputation of missing values
  for incomplete pattern classification, Pattern Recognition 52 (2016) 85--95.

\bibitem{Liu2018Change}
Z.~G. Liu, L.~Gang, G.~Mercier, H.~You, P.~Quan, Change detection in
  heterogenous remote sensing images via homogeneous pixel transformation, IEEE
  Transactions on Image Processing PP~(99) (2018) 1--1.

\bibitem{dutta2015uncertainty}
P.~Dutta, Uncertainty modeling in risk assessment based on {Dempster--Shafer}
  theory of evidence with generalized fuzzy focal elements, Fuzzy Information
  and Engineering 7~(1) (2015) 15--30.

\bibitem{zhengxianglin2017}
X.~Zheng, Y.~Deng, Dependence assessment in human reliability analysis based on
  evidence credibility decay model and {IOWA} operator, Annals of Nuclear
  Energy (2017) accepted.

\bibitem{zhang2017improved}
L.~Zhang, L.~Ding, X.~Wu, M.~J. Skibniewski, An improved {Dempster--Shafer}
  approach to construction safety risk perception, Knowledge-Based Systems.

\bibitem{liuDEMATEL2017}
T.~Liu, Y.~Deng, F.~Chan, Evidential supplier selection based on {DEMATEL} and
  game theory, International Journal of Fuzzy Systems 20~(4) (2018) 1321--1333.

\bibitem{Jiang2017FMEA}
W.~Jiang, C.~Xie, M.~Zhuang, Y.~Tang, Failure mode and effects analysis based
  on a novel fuzzy evidential method, Applied Soft Computing 57 (2017)
  672--683.

\bibitem{fan2006fault}
X.~Fan, M.~J. Zuo, Fault diagnosis of machines based on {D--S} evidence theory.
  {Part 1: D--S} evidence theory and its improvement, Pattern Recognition
  Letters 27~(5) (2006) 366--376.

\bibitem{Du2016FMEA}
Y.~Du, X.~Lu, X.~Su, Y.~Hu, Y.~Deng, New failure mode and effects analysis: An
  evidential downscaling method, Quality and Reliability Engineering
  International 32~(2) (2016) 737--746.

\bibitem{zheng2017evaluation}
H.~Zheng, Y.~Deng, {Evaluation method based on fuzzy relations between
  Dempster-Shafer belief structure}, {International Journal of Intelligent
  Systems} ({2017}) Article ID: INT21956\href
  {http://dx.doi.org/{10.1002/int.21956}} {\path{doi:{10.1002/int.21956}}}.

\bibitem{Liu2017Classifier}
Z.~Liu, Q.~Pan, J.~Dezert, J.~W. Han, Y.~He, Classifier fusion with contextual
  reliability evaluation., IEEE Transactions on Cybernetics PP~(99) (2017)
  1--14.

\bibitem{kangbingyi2017}
B.~Kang, G.~Chhipi-Shrestha, Y.~Deng, J.~Mori, K.~Hewage, R.~Sadiq,
  {Development of a predictive model for Clostridium difficile infection
  incidence in hospitals using Gaussian mixture model and Dempster-Shafer
  theroy}, Stochastic Environmental Research and Risk Assessment~(1) (2017)
  1--16.

\bibitem{Jiang2016CAIE}
W.~Jiang, B.~Wei, J.~Zhan, C.~Xie, D.~Zhou, A visibility graph power averaging
  aggregation operator: A methodology based on network analysis, Computers \&
  Industrial Engineering 101 (2016) 260--268.

\bibitem{dong2017location}
Y.~Dong, J.~Wang, F.~Chen, Y.~Hu, Y.~Deng, Location of facility based on
  simulated annealing and {``ZKW''} algorithms, Mathematical Problems in
  Engineering 2017 (2017) Article ID 4628501.

\bibitem{Den2014Optimal}
T.~Den{\oe}ux, N.~E. Zoghby, V.~Cherfaoui, A.~Jouglet, Optimal object
  association in the {Dempster-Shafer} framework, IEEE Transactions on
  Cybernetics 44~(12) (2014) 2521--2531.

\bibitem{zheng2017fuzzy}
H.~Zheng, Y.~Deng, Y.~Hu, Fuzzy evidential influence diagram and its evaluation
  algorithm, Knowledge-Based Systems {131} (2017) {28--45}.

\bibitem{zadeh1986simple}
L.~A. Zadeh, A simple view of the {Dempster--Shafer} theory of evidence and its
  implication for the rule of combination, AI Magazine 7~(2) (1986) 85.

\bibitem{lefevre2002belief}
E.~Lefevre, O.~Colot, P.~Vannoorenberghe, Belief function combination and
  conflict management, Information Fusion 3~(2) (2002) 149--162.

\bibitem{Jiang2017Ordered}
W.~Jiang, B.~Wei, Y.~Tang, D.~Zhou, Ordered visibility graph average
  aggregation operator: An application in produced water management, Chaos: An
  Interdisciplinary Journal of Nonlinear Science 27~(2) (2017) 023117.

\bibitem{ma2015belief}
J.~Ma, W.~Liu, S.~Benferhat, A belief revision framework for revising epistemic
  states with partial epistemic states, International Journal of Approximate
  Reasoning 59 (2015) 20--40.

\bibitem{smets1990combination}
P.~Smets, The combination of evidence in the transferable belief model, IEEE
  Transactions on Pattern Analysis and Machine Intelligence 12~(5) (1990)
  447--458.

\bibitem{dubois1988representation}
D.~Dubois, H.~Prade, Representation and combination of uncertainty with belief
  functions and possibility measures, Computational Intelligence 4~(3) (1988)
  244--264.

\bibitem{yager1987dempster}
R.~R. Yager, On the {Dempster-Shafer} framework and new combination rules,
  Information Sciences 41~(2) (1987) 93--137.

\bibitem{murphy2000combining}
C.~K. Murphy, Combining belief functions when evidence conflicts, Decision
  Support Systems 29~(1) (2000) 1--9.

\bibitem{zhang2014novel}
Z.~Zhang, T.~Liu, D.~Chen, W.~Zhang, Novel algorithm for identifying and fusing
  conflicting data in wireless sensor networks, Sensors 14~(6) (2014)
  9562--9581.

\bibitem{Pothos2009A}
E.~M. Pothos, J.~R. Busemeyer, A quantum probability explanation for violations
  of 'rational' decision theory, Proceedings of the Royal Society of London
  Biological Sciences 276~(1665) (2009) 2171--2178.

\bibitem{busemeyer2012quantum}
J.~R. Busemeyer, P.~D. Bruza, Quantum models of cognition and decision,
  Cambridge University Press, 2012.

\bibitem{Bruza2015Quantum}
Bruza, D.~Peter, Wang, Zheng, Busemeyer, R.~Jerome, Quantum cognition: a new
  theoretical approach to psychology, Trends in Cognitive Sciences 19~(7)
  (2015) 383--93.

\bibitem{ablowitz2003complex}
M.~J. Ablowitz, A.~S. Fokas, Complex variables: introduction and applications,
  Cambridge University Press, 2003.

\bibitem{deng2017mess}
Y.~Deng, Meta mass function (2017) vixra preprint vixra:1708.0065v1.

\bibitem{Busemeyer2009Empirical}
J.~R. Busemeyer, Z.~Wang, A.~Lambert-Mogiliansky, Empirical comparison of
  {Markov} and quantum models of decision making, Journal of Mathematical
  Psychology 53~(5) (2009) 423--433.

\bibitem{wang2016interference}
Z.~Wang, J.~R. Busemeyer, Interference effects of categorization on decision
  making, Cognition 150 (2016) 133--149.

\bibitem{Townsend2000Exploring}
J.~T. Townsend, K.~M. Silva, J.~Spencersmith, M.~J. Wenger, Exploring the
  relations between categorization and decision making with regard to realistic
  face stimuli, Pragmatics \& Cognition 8~(1) (2000) 83--105.

\bibitem{He2017Anevidential}
Z.~He, W.~Jiang, An evidential {Markov} decision making model, Proceedings of
  the Royal Society of London Biological Sciences (2017) arXiv preprint
  arXiv:1705.06578.

\end{thebibliography}

\end{spacing}
\end{document}